\documentclass[sigconf]{acmart}

\usepackage{booktabs} 
\usepackage{graphicx}
\usepackage{amsmath}

\setcopyright{none}
\renewcommand\footnotetextcopyrightpermission[1]{}
\setcitestyle{square}

\begin{document}

\title{Deep-VFX: Deep Action Recognition Driven VFX for Short Video}

\author{Ao Luo}
\affiliation{%
  \institution{UESTC}}
\email{2015060501020@std.uestc.edu.cn}

\author{Ning Xie}
\affiliation{%
  \institution{UESTC}}
\email{xiening@uestc.edu.cn}

\author{Zhijia Tao}
\affiliation{%
  \institution{UESTC}}
\email{}

\author{Feng Jiang}
\affiliation{%
  \institution{UESTC}}
\email{}

\renewcommand{\shortauthors}{DeJohnette, Rowland-Smith, Badeeri, and Foyt}

\keywords{Motion Capture, LSTM, Skeleton-based Action Recognition, VFX}

\begin{teaserfigure}
  \centering
  \includegraphics[width=\linewidth]{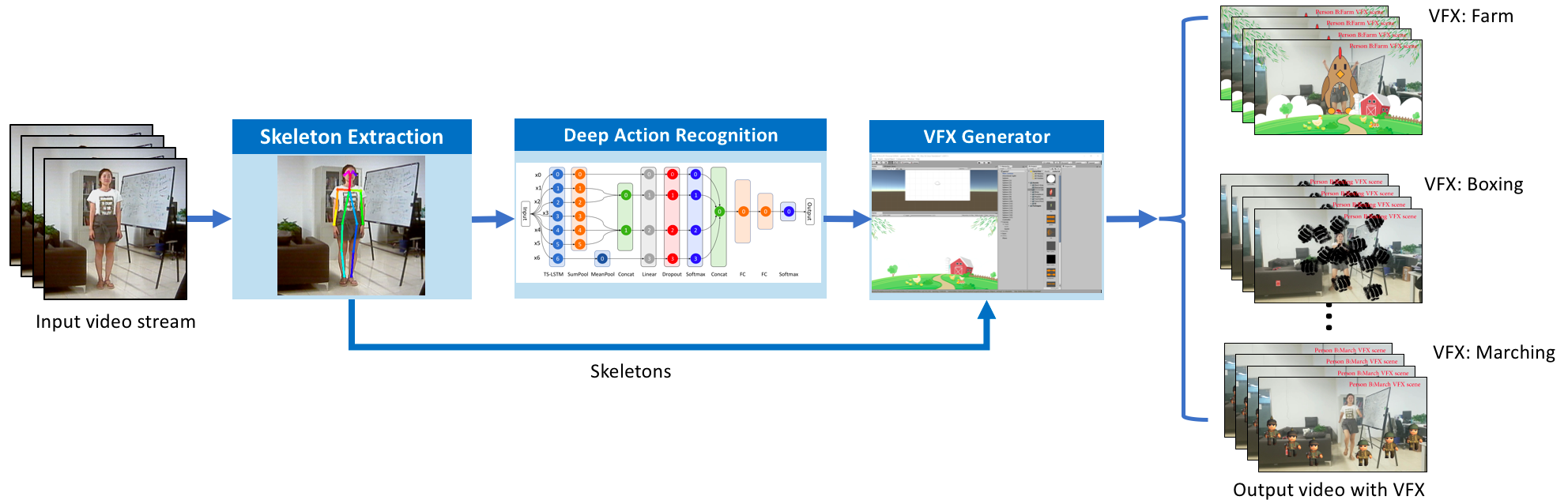}
  \caption{System architecture of Deep-VFX. There are three key modules, which are the skeleton extraction, 
  the deep action recognition, and the VFX generation. By using this motion-driven AI system, users enable to generate VFX 
  short video in the personal style.}
  \label{fig:teaser}
\end{teaserfigure}

\maketitle

\section{Introduction}
Human motion is a key function to communicate information. In the application, short-form mobile video is so popular all over the world such as Tik Tok. Recently, users would like to add more VFX so as to pursue creativity and personality. Many special effects therefore are made for the short video platform. To do so, the common manner is to create many templates of VFX. However, these preformed templates are not easy-to-use. Because the timing and the content are fixed. Users can not edit the factors as their desired. The only thing they can do is to do tedious attempt to grasp the timing and rhythm of the non-modifiable templates. 

This paper aims to provide an user-centered VFX generation method. We propose the LSTM-based action recognition which can identify users' intention by making actions. Together with the human body key point tracking, our system {\it Deep-VFX} can help users to generate the VFX short video according to the meaningful gestures with users' specific personal rhythm. In details, as illustrated in Figure~\ref{fig:teaser}, the proposed Deep-VFX system are composited by three key modules. The skeleton extraction module works to calculate the key points of the user's body (see Section~\ref{sec:skeleton_extraction}). We propose a novel form of intensive TS-LSTM ({\it iTS-LSTM}) to find out the user's intention in the deep action recognition module (see Section~\ref{sec:itslstm}). The VFX generator works to render the short video with the animation assets as post processing.The experimental results demonstrate that our AI-based method achieves the purpose for the motion-driven personal VFX generation of the short video more easily and efficiently.

\section{Human Body Skeleton Extraction}
\label{sec:skeleton_extraction}
In VFX generation, the special effects highly reply on the skeleton of human body~\cite{wei2016cpm}, face and hand~\cite{simon2017hand}. The marker-less motion capture method~\cite{Cao2017Realtime} called {\it OpenPose} is applied in our proposed deep-VFX system. However, the key points dropping in the frames is the vital issue in the real time VFX task. After our well study, the problem comes from the noise caused by the camera hardware. Therefore, we propose the method to restrain both brightness instability and salt-and-pepper noise among continuous frames so as to to guarantee the fluency and stable of the key points in frames. In order to guarantee the real-time performance, all operations of video processing are implemented in CUDA to speed up.

\section{Intensive TS-LSTM Method for Skeleton-based Action Recognition}
\label{sec:itslstm}

\begin{figure}[ht]
  \centering
  \includegraphics[width=.8\linewidth]{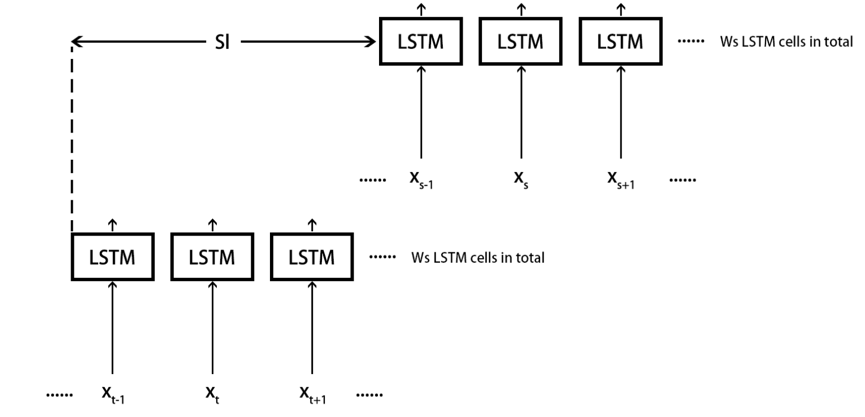}
  \caption{Conceptual diagram of the TS-LSTM module.}
  \label{tslstm} 
\end{figure}

\begin{figure}[ht]
  \centering
  \includegraphics[width=\linewidth]{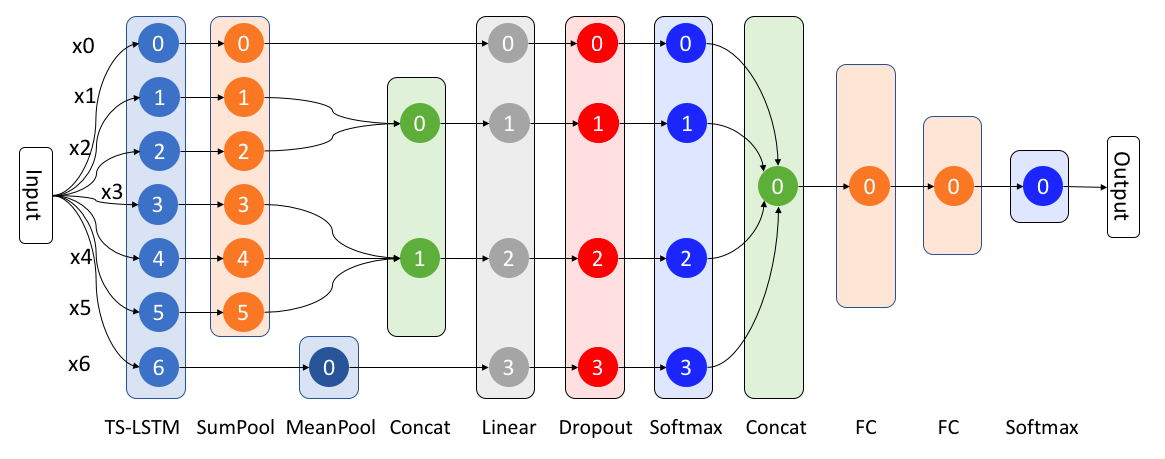}
  \caption{The architecture of iTS-LSTM with the short-term, medium-term and original 
  modules.}
  \label{fig:detailofitslstm} 
\end{figure}

LSTM network is suitable to process the sequence task. The most successful domain is natural language processing (NLP). It has excellent ability for learning long and short language sentences~\cite{Graves1997Long}. The main topic of this paper is skeleton-based action recognition which is similar to NLP in the aspect of the vector representation. But different from NLP, the action recognition is sensitive to the rhythm variety of actions. Therefore, we use the recommended structure, Temporal Sliding LSTM (TS-LSTM)~\cite{Lee2017Ensemble}. 


By taking different parameters in the input layer of TS-LSTM structure, we trap the memory of different temporal terms for the actions. By doing this, we solve the action rhythm various problem. The base of TS-LSTM cell is shown as Figure~ref{tslstm}. It has a sliding window, this sliding window consists of $W_l$ (Window length) LSTM cells. TS-LSTM cell moves on the input sequence, it jumps $\mbox{TS}_l$ (Temporal Sliding length) frames of the input sequence for each time. The whole network comprises of 7 TS-LSTM cells in total, each cell has different inputs and effects. We improve the original structure of TS-LSTM by adding the component with two neighbouring full connection layers and softmax layer as illustrated in Figure~\ref{fig:detailofitslstm}. This new form enhances the capability of the neural network in learning multiple features can be fully exploited. We call it as intensive TS-LSTM (iTS-LSTM).

%

\section{Experiment}
In this section, we experimentally evaluate the performance the efficiency of our proposed method through the comparison with other related algorithms. Then, we will show the results of acton driven VFX for short videos in practical.

\begin{table}[h]
\centering
\caption{Parameters of iTS-LSTM. The concrete parameters for each iTS-LSTM network. The Hidden Size (Hs) is the node number of hidden layer in every LSTM network. }
\label{tab:ts_lstm_parameter}
\begin{tabular}{|c|c|c|c|c|c|}
\hline
 & $H_s$ & $D_l$ & $W_l$ & $\mbox{TS}_l$ & LN \\\hline
$\mbox{iTS-LSTM}_0$ & 256 & 1 & 5 & 5 & 128 \\\hline
$\mbox{iTS-LSTM}_1$ & 256 & 1 & 11 & 11 & 64\\\hline
$\mbox{iTS-LSTM}_2$ & 256 & 5 & 9 & 9 & - \\\hline
$\mbox{iTS-LSTM}_3$ & 256 & 1 & 23 & - & 32\\\hline
$\mbox{iTS-LSTM}_4$ & 256 & 5 & 19 & - & -\\\hline
$\mbox{iTS-LSTM}_5$ & 256 & 10 & 14 & - & -\\\hline
$\mbox{iTS-LSTM}_6$ & 256 & 0 & 12 & 12 & 64\\\hline
\end{tabular}
\end{table}

We introduce the configuration of hyper-parameters for the iTS-LSTM network. There are 24 frames in an input skeleton sequence, each frame contains 18 skeleton 2D points $(x, y)$, 36 float numbers in total. The optimizer we use is Adam, the original learning rate is $1e-4$, to ensure the model gradually approach the global optimal result without trapping in a seriously wrong local result. We set the dropout proportion to 0.2 to keep from overfitting. As we seen, the window sizes are same with sliding length, the reason is that, with the network kept as most information as it could, we need to improve the real-time ability of the network, so we did not add the overlap at the data input part. After the last concatation, there are 2 fully connected layers utilized for output. The node number for fc1 is 72, ${fc}_2$ is 18. Finally, ${fc}_2$ output to softmax layer, do classification for 4 classes. In conclusion, the accuracy of proposed iTS-LSTM is the highest in Table~\ref{tab:result_action_recognition}. It shows that iTS-LSTM has better performance with lower risk of overfitting. 

\begin{table}[h]
\centering
\caption{The comparison on the accuracy and loss.}
\label{tab:result_action_recognition}
\begin{tabular}{|c|c|c|}
\hline
                 & ACC.(\%) & Loss (Entropy) \\\hline
iTS-LSTMs & $\mathbf{95.30}$ & 0.0748 \\\hline
TS-LSTMs (with original data) & $94.80$ & $\mathbf{0.0472}$ \\\hline
TS-LSTMs (without original data) & $94.36$ & 0.1142 \\\hline
Double LSTMs & 93.12 & 0.0920 \\\hline
Single LSTM & 94.09 & 0.0682 \\\hline
\end{tabular}
\end{table}

\section{Conclusion}
In this paper, we proposed the user-centered VFX generation system. It has three main contributions: (1) The skeleton extraction module works to calculate the key points of the user's body; (2) The novel form of intensive TS-LSTM ({\it iTS-LSTM}) to find out the user's intention in the deep action recognition module; (3) The VFX generator works to render the short video with the animation assets as post processing.  The experimental results demonstrate that our AI-based method achieves the purpose for the motion-driven personal VFX generation of the short video more easily and efficiently. In future work, we plan to create more 3D VFX assets of the props and special effects instead of the current 2D effects.


\bibliographystyle{ACM-Reference-Format}
\bibliography{actiondrivenvfx}

\end{document}